\documentclass{bmvc2k}

\title{GN-FR: Generalizable Neural Radiance Fields for Flare Removal}

\addauthor{Gopi Raju Matta${^*}$}{ee17d021@smail.iitm.ac.in}{1}
\addauthor{Rahul Siddartha}{ee22m015@smail.iitm.ac.in}{1}
\addauthor{Rongali Simhachala Venkata Girish}{girishrongali9@gmail.com }{1}
\addauthor{Sumit Sharma}{ee20d042@smail.iitm.ac.in}{1}
\addauthor{Kaushik Mitra}{kmitra@ee.iitm.ac.in}{1}

\addinstitution{
 Computational Imaging Lab,\\
 Dept. of Electrical Engineering,\\
 Indian Institute of Technology Madras,\\
 Chennai, India
}

\runninghead{MATTA, Siddartha, Girish, Sharma, Mitra}{
GN-FR
}

\usepackage{amsmath}
\usepackage{amssymb}
\usepackage{amsfonts}
\usepackage{amstext}
\usepackage{amsthm}
\usepackage{caption}
\usepackage{subcaption}
\usepackage{amsmath,amssymb,graphicx}
\usepackage{hyperref}
\usepackage{booktabs}
\usepackage{multirow}
\usepackage{mathtools}
\usepackage{cleveref}
\usepackage{xcolor,soul}
\usepackage{colortbl}
\usepackage[export]{adjustbox}
\usepackage{tikz}
\usetikzlibrary{spy}
\usepackage{graphicx}
\usepackage{diagbox}
\usepackage{makecell}
\usepackage{amssymb}
\usepackage{enumitem}
\usepackage{booktabs}
\usepackage{pifont}
\newcommand{\cmark}{\ding{51}}%
\newcommand{\xmark}{\ding{55}}%

\begin{document}

\maketitle
\let\thefootnote\relax\footnotetext{* Corresponding author}
\begin{abstract}
      Flare, an optical phenomenon resulting from unwanted scattering and reflections within a lens system, presents a significant challenge in imaging. The diverse patterns of flares—such as halos, streaks, color bleeding, and haze—complicate the flare removal process. Existing traditional and learning-based methods have exhibited limited efficacy
due to their reliance on single-image approaches, where flare removal is highly ill-posed. We address this by framing flare removal as a multi-view image problem, taking advantage of the view-dependent nature of flare artifacts. This approach leverages information from neighboring views to recover details obscured by flare in individual images. Our proposed framework, GN-FR (Generalizable Neural Radiance Fields for Flare Removal), can render flare-free views from a sparse set of input images affected by lens flare and generalizes across different scenes in an unsupervised manner. GN-FR incorporates several modules within the Generalizable NeRF Transformer (GNT) framework: Flare-occupancy Mask Generation (FMG), View Sampler (VS), and Point Sampler (PS). To overcome the impracticality of capturing both flare-corrupted and flare-free data, we introduce a masking loss function that utilizes mask information in an unsupervised setting. Additionally, we present the first-of-its-kind 3D multi-view flare dataset, comprising 17 real flare scenes with 782 images, 80 real flare patterns, and their corresponding annotated flare-occupancy masks. To our knowledge, this is the first work to address flare removal within a Neural Radiance Fields (NeRF) framework.

\end{abstract}
\section{Introduction}

The existence of flare artifacts can degrade image quality, potentially obscuring crucial object or scene details.  Variations in lens characteristics, light source placement, and the camera's orientation towards the light source can result in diverse flare types. This diversity makes flare removal an ill-posed problem. Recent works \cite{wu2021train,dai2023flare7k++,dai2023nighttime,zhang2023ff} attempted to remove flare by utilizing the flare mask as a guiding cue in conjunction with paired data by using Unet and transformer-based architectures. Qiao \textit{et al.}\cite{qiao2021light} utilized adversarial networks to learn from both flare and flare-free unpaired data by acknowledging the position of flare and light source. Lately, kotp \textit{et al.}\cite{kotp2024flare} leveraged depth information as a guide to remove flare. However, none of the aforementioned papers exploited the directional dependence of flare. We propose to utilize multi-view settings for flare removal, leveraging information lost due to flare in the current view from adjacent views. With the current framework, we can achieve novel view synthesis from images corrupted by flare and simultaneously eliminate the flare artifacts.
\vspace{-1em}
\begin{figure*}[!ht]
\centering
  \includegraphics[width=0.9\textwidth]{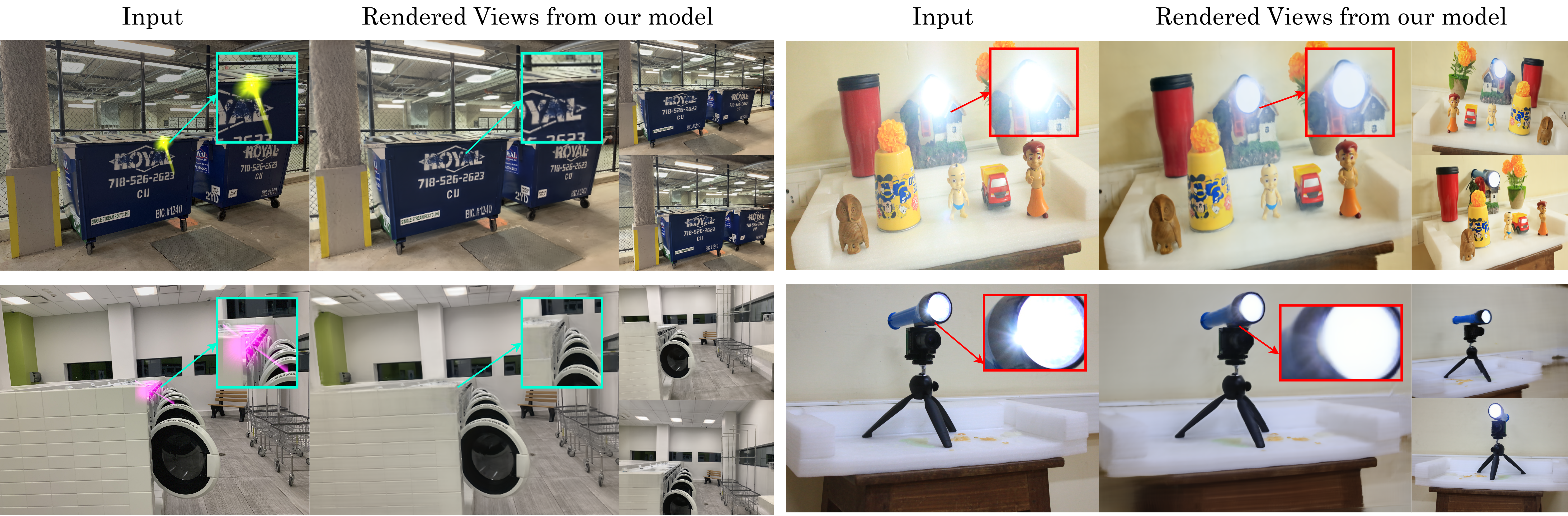}
  \caption{\textbf{Our method:} GN-FR is built on significant modifications to the GNT framework with flare masking loss, an effective view sampler and a Point Sampler. Our method consistently removes flare across the views on synthetic flare-imposed scenes and real scenes with flare.}
  \label{fig:teaser}
\end{figure*}

Neural scene representations such as NeRF\cite{mildenhall2020nerf} can render realistic novel views, but they are limited by a tedious optimization process for each new scene and cannot generalize to new scenes. Recent works on generalizable neural rendering like IBRNet\cite{wang2021ibrnet}, PixelNerf\cite{yu2021pixelnerf}, NeuRay\cite{liu2022neural} address these limitations by synthesizing a generalizable 3D representation through aggregation of image features extracted from seen views.
We leverage GNT\cite{gnt}, a recent powerful method that utilizes transformers as aggregation functions to extract features from source views and to predict the target pixel color, eliminating the usage of the volumetric rendering equation. This makes GNT a reliable framework for flare removal alongside generating novel views. However, vanilla GNT cannot remove the flare as the network cannot learn to identify the flare region without sufficient cues. By leveraging flare occupancy masks as cues, we direct the network's attention to the flare-affected regions, enabling it to learn how to remove flare and recover lost information from neighboring views effectively. This underscores our central concept that information obscured in one view can be extracted from adjacent views. 
 
 To this end, we propose our model GN-FR: Generalizable Neural Radiance Fields for Flare Removal, which is built on effective modifications to the GNT framework. Our model includes three sub-modules:
 
 1) \textbf{Flare-occupancy Mask Generation network} (FMG): For each input view, it generates a binary mask to detect the regions w/ and w/o flare. 
 
 2) \textbf{View Sampler} (VS): Samples the views that are minimally affected by flare using the flare occupancy mask as a cue. 
 
 3) \textbf{Point Sampler} (PS): An effective attention masking-based sampling mechanism in the view transformer to guide the network to attend only to points that are valid in the flare occupancy mask. 
 
 As ground truth is not available for real scenes, we propose \textbf{Masking loss}, this will only attend to the pixels where there is no flare while calculating loss. Hence it ensures the rendering of the regions where the flare is prominent by extracting information from corresponding nearby views' regions where there is no flare.
 
In summary, our contributions include:
\begin{itemize}
    \item A generalizable neural rendering network for flare removal that can render novel views and effectively remove the flare in complex scenarios.
    \item Masking loss, Flare-occupancy Mask Generator (FMG), View Sampler (VS) and Point Sampler (PS) modules to intelligently sample only flare-free regions in the source views for target view rendering. 
    \item A first-of-its-kind 3D flare dataset of 17 scenes, together with real flare pattern dataset of 80 different flare patterns of real flare to enable our network to detect flare affected regions well and then to effectively de-flare them. 
\end{itemize}
To the best of our knowledge, this is the first work that does flare removal and also produces novel views in a single framework.

\vspace{-0.75em}
\section{Related Work}
\vspace{-0.75em}

\subsection{Flare Removal}
Flare removal remained a challenging problem for a longer period due to its inherent complexity and diverse reasons causing flare. Despite these challenges, recent traditional \cite{zhou2023improving} and deep learning \cite{dai2023flare7k++,dai2023nighttime} based methods have shown decent progress. Wu \textit{et al.}\cite{wu2021train} proposed Unet-based flare removal using paired data and flare occupancy mask information, where the mask information is used to blend the input light source back into the prediction where the light source is removed during training.Qiao \textit{et al.}\cite{qiao2021light} utilized the position of the light source and flare to remove the flare effectively.
\cite{dai2023flare7k++,dai2023nighttime} used Uformer\cite{wang2022uformer} to remove flare. 
While the above methods primarily focus on single-image flare removal, they overlook the potential benefits of multi-view image flare removal. In multi-view approaches, information lost due to flare in one image can be potentially recovered from neighboring views, enhancing the overall quality of the reconstructed image.

\vspace{-0.75em}
\subsection{Novel View Synthesis}
NeRF introduced by \cite{mildenhall2020nerf} synthesizes consistent and photorealistic novel views by fitting each scene as a continuous 5D radiance field parameterized by an MLP(Multi-layer Perceptron). Several efforts have been made recently to improve efficiency and quality \cite{verbin2022ref, Barron_2021_ICCV,hu2022efficientnerf,chen2021mvsnerf} of NeRF. Some works have introduced lighting and reflection modeling \cite{chen2021nerv, verbin2022ref} to NeRF. Recently, Mildenhall \textit{et al.} proposed RawNeRF\cite{mildenhall2022nerf}, which can yield novel HDR views by training directly on linear RAW images. 
\vspace{-0.75em}
\subsection{Generalizable NeRF}
The original NeRF\cite{mildenhall2020nerf} approach is limited to training a neural network for individual scenes, requiring optimization from scratch for each new scene without prior knowledge. Generalizable neural rendering methods overcome this constraint by training on multiple scenes, allowing the network to develop a broader understanding of synthesizing novel views by leveraging source observations. Earlier methods acquire RGB$\sigma$ of each point by weighted summing of the image features of its 2D projections\cite{yu2021pixelnerf} and through cost volume induced by MVSNet\cite{chen2021mvsnerf}. To enhance generalization capabilities and rendering quality, recent approaches have incorporated transformer-based architectures \cite{dosovitskiy2020image} to aggregate features from source images \cite{liu2022neural,wang2022generalizable}. IBRNet\cite{wang2021ibrnet}uses MLP conditioned on feature vectors extracted from the source images to predict color and radiance values, which are aggregated using volumetric rendering and a ray transformer to compute densities along
the camera ray. GNT\cite{gnt} uses transformers for the entire pipeline. However, the above methods cannot remove flare while generating novel views.

\vspace{-0.75em}
\section{Preliminary}
\paragraph{Generalizable Neural Radiance Fields:}
Vanilla NeRF\cite{mildenhall2020nerf} represents the novel views using volumetric rendering, which is time-consuming and produces artifacts while rendering reflective and metal surfaces. GNT\cite{gnt} overcomes these issues by
considering the problem of novel view synthesis as a two-stage information process: the multi-view image feature fusion, followed by the sampling-based rendering integration. It is composed of (i) \textit{view transformer} to aggregate pixel-aligned image features from corresponding epipolar lines to predict coordinate-wise features, (ii) \textit{ray transformer} to compose coordinate-wise point features along a traced ray via attention mechanism. More formally, the entire operation can be summarized as follows:

\begin{footnotesize}
\begin{align}
    \begin{split}
        \mathcal{F}(\mathbf{x}, \mathbf{\theta}) = \operatorname{View-Transformer} (&\mathbf{F}_1(\Pi_1(\mathbf{x}), \mathbf{\theta}), \cdots, \mathbf{F}_N(\Pi_N(\mathbf{x}), \mathbf{\theta})    
    \end{split}
\end{align}
\end{footnotesize}

where $\operatorname{View-Transformer}\left(\cdot\right)$ is a transformer encoder, $\Pi_i\left(\mathbf{x}\right)$ projects position $\mathbf{x} \in \mathbb{R}^3$ onto the $i^{th}$ image plane by applying projection matrix, $\Pi$ onto $\mathbf{x}$ and $\mathbf{F}_i(\mathbf{z}, \mathbf{\theta}) \in \mathbb{R}^d$ computes the feature vector at position $\mathbf{z} \in \mathbb{R}^2$ via bi-linear interpolation on the feature grids. The multi-view aggregated point features are fed into the ray transformer, and the output from the ray transformer is fed into a view transformer and this process is repeated where the view transformer and ray transformer are stacked alternatively. Then the features from the last ray transformer are pooled to extract a single ray feature to predict the target pixel color. 

\begin{footnotesize}
\begin{align}
    \begin{split}
        \mathbf{C}(\mathbf{r}) = \operatorname{MLP} \circ \operatorname{Mean} \circ \operatorname{Ray-Transformer}(&\mathcal{F}(\mathbf{o} + t_1 \mathbf{d}, \mathbf{\theta}), \cdots, \mathcal{F}(\mathbf{o} + t_M \mathbf{d}, \mathbf{\theta}))
    \end{split}
\end{align}
\end{footnotesize}

where $t_1, \cdots, t_M$ are uniformly sampled  between near and far plane, $\operatorname{Ray-Transformer}$ is a standard transformer encoder.

\vspace{-0.75em}
\section{Method: Generalizable Neural Radiance Fields
for Flare Removal}
\label{sec:methods}

\begin{figure*}[!ht]
\centering
  \includegraphics[width=0.95\textwidth]{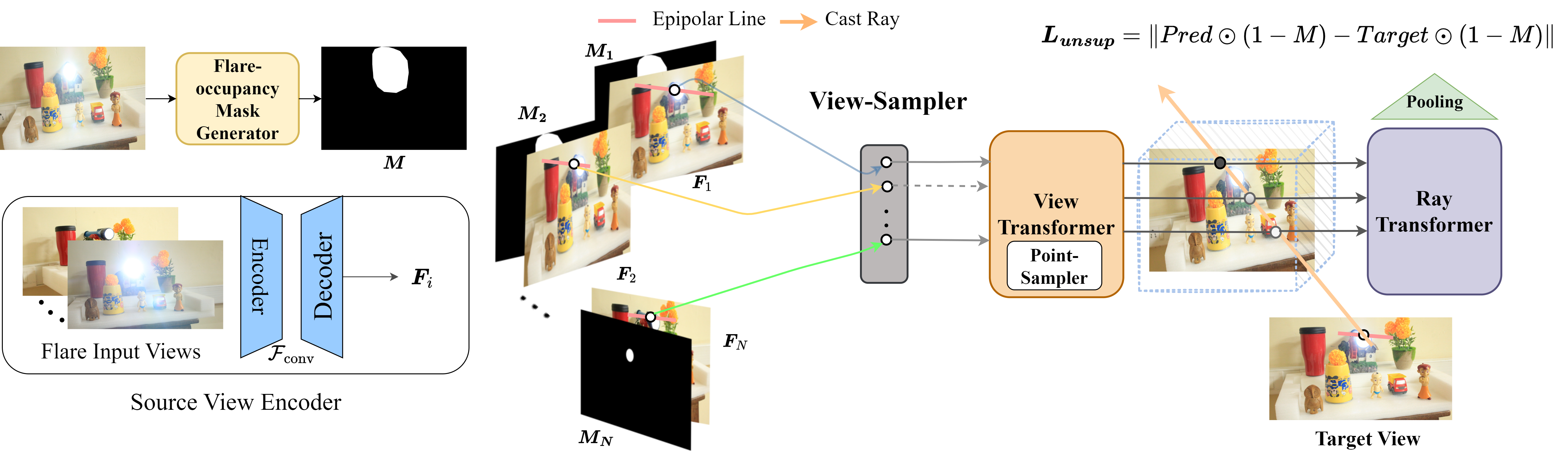}
  \caption{\textbf{Overview of GN-FR:} 1) FMG: Flare-occupancy Mask Generation network: It generates a binary mask for each input view to detect the regions w/ and w/o flare. 2) View Sampler: Samples the views minimally affected by flare using the flare occupancy mask as a cue. 3) Point Sampler: An attention masking-based sampling mechanism in the view transformer to guide the network for better flare removal.}
  \label{fig:method}
\end{figure*}

\textbf{Overview:} We introduce GN-FR to perform generalizable novel view synthesis on flare-corrupted images. Given a set of $N$ input views with known camera parameters $\{ ({I}_i  \in \mathbb{R}^{H \times W \times 3}, {P}_i \in \mathbb{R}^{3 \times 4}) \}_{i=1}^{N}$, corrupted with flare, our goal is to synthesize novel views from arbitrary camera positions and also generalize to new scenes.

Our pipeline has three stages, as shown in Fig.~\ref{fig:method}. Specifically, given a set of $N$ flare input images, GN-FR utilizes a UNet with residual connection to encode each image to a feature map, expecting it to extract shading and local/global complex light transport information via its multi-scale architecture.
We employ \textbf{\textit{FMG}}( Flare-occupancy Mask Generator) (Sec.~\ref{sec:fmg}) to generate a mask (\textbf{\textit{M}}) that contains zero values wherever there is no flare. 
Our FMG is capable of discerning flare regions effectively, excluding areas corresponding to the light source. This mask is used as a cue to guide the flare removal at regular stages of the network.

Following GNT\cite{gnt}, our method projects \textbf{\textit{x}} (position (x,y,z)) onto every source image and interpolates the feature vector on the image plane to obtain the feature representation at that position \textbf{\textit{x}}. To render a target view, vanilla GNT\cite{gnt} selects 8-10 nearby source views based on their relative distance. We modify this via our \textbf{\textit{View Sampler}} (Sec.~\ref{sec:viewandpoint}), which samples only those nearby views based on the flare occupancy mask information; typically, the source views where the percentage of flare is relatively less from other source views.
We then use a \textbf{\textit{Point Sampler}} (Sec.~\ref{sec:viewandpoint}) imbibed View transformer to aggregate all the feature vectors by masking the points affected by flare while computing attention. Such aggregated features are passed through the ray transformer (Eq.2) to render the final color. The network is trained end-to-end using the \textbf{\textit{Masking loss}} (Sec.~\ref{sec:loss}) in an unsupervised fashion.

\vspace{-0.75em}
\subsection{Flare-occupancy Mask Generation}
\label{sec:fmg}

\begin{figure}[h]
    \centering
    \begin{tabular}{cc}
        \begin{minipage}{0.318\textwidth}
            \centering
            \includegraphics[width=\textwidth]{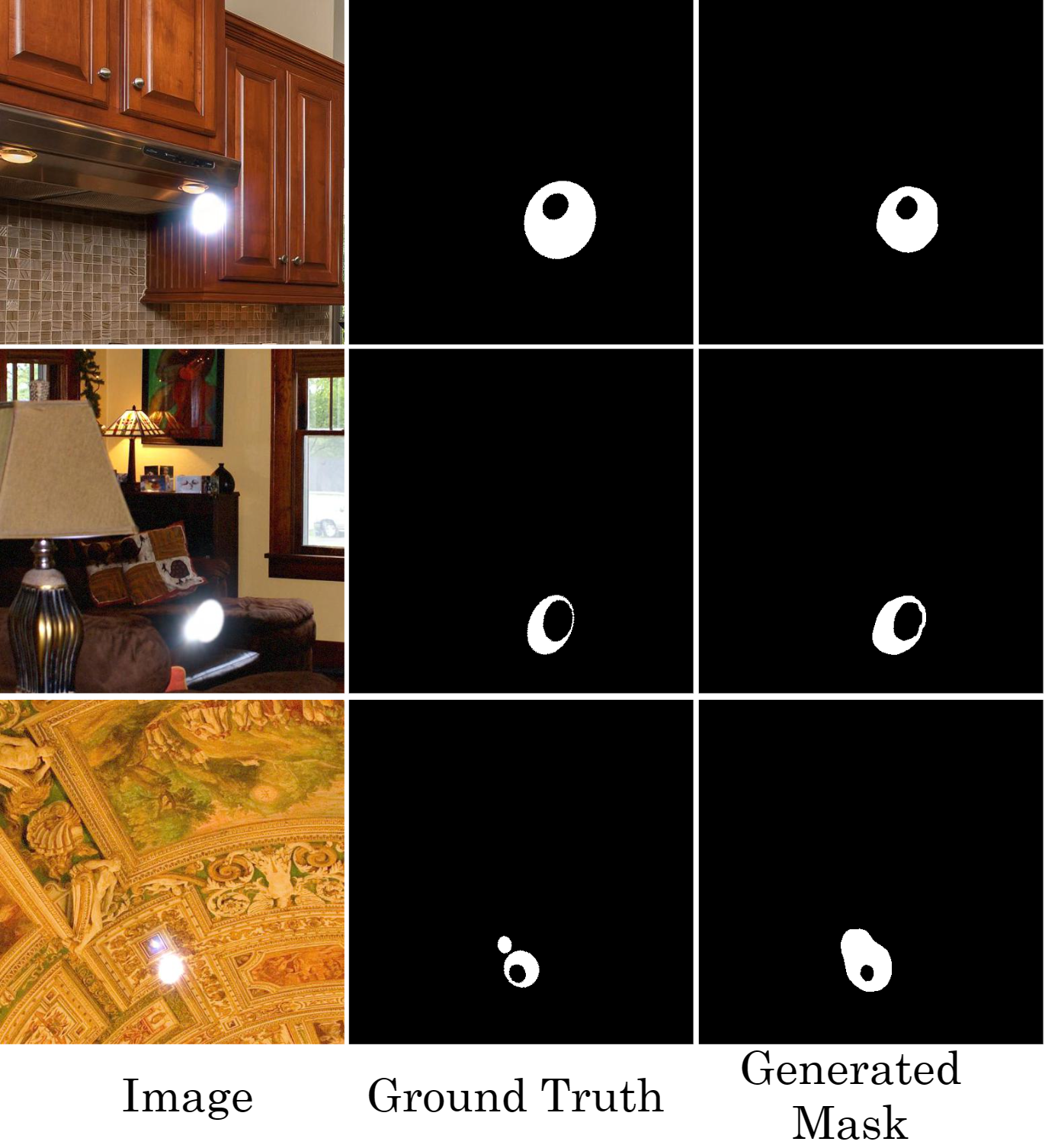}
            \caption*{(a) Synthetic Dataset}
            \label{fig:synthetic}
        \end{minipage} &
        \begin{minipage}{0.516\textwidth}
            \centering
            \includegraphics[width=\textwidth]{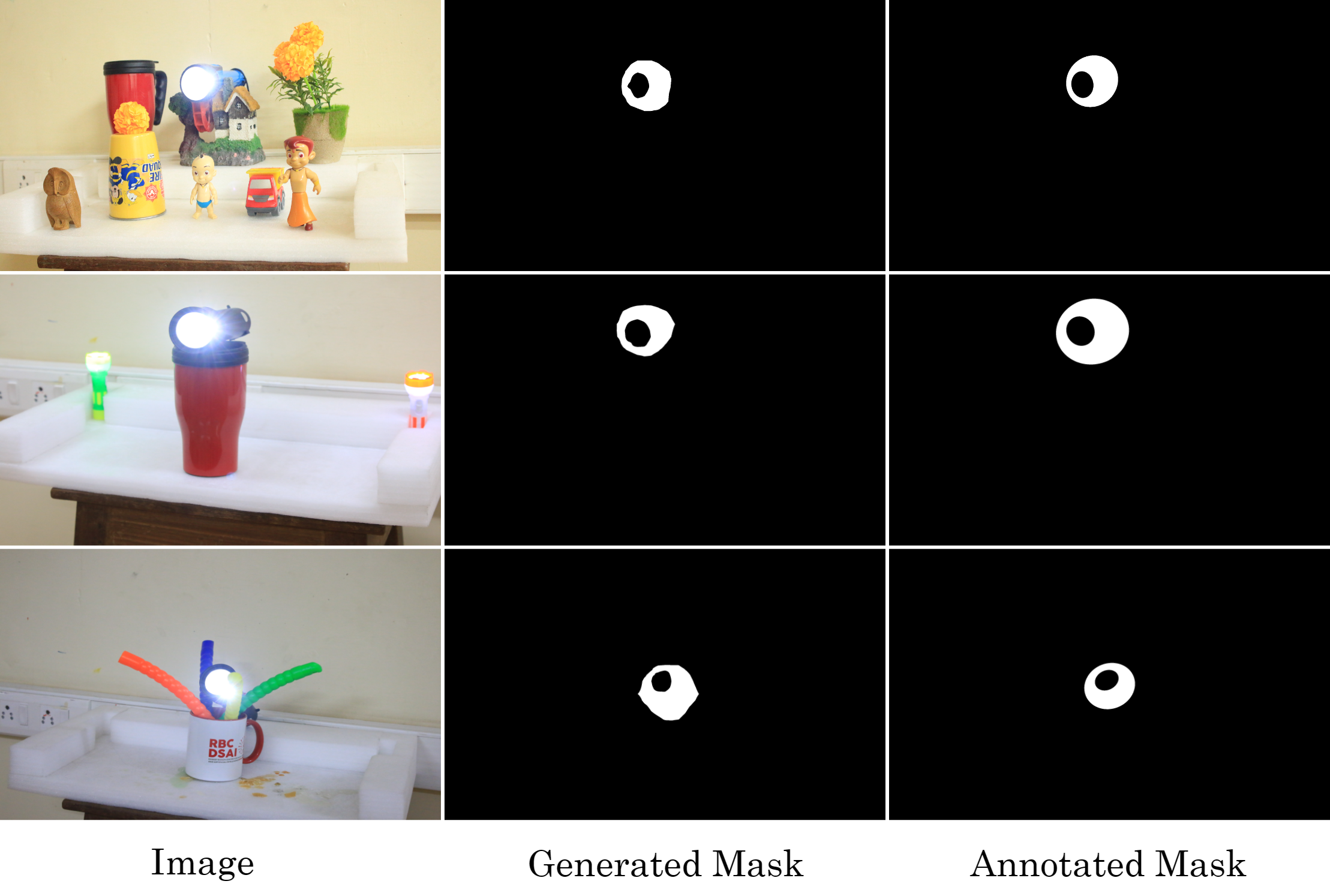}
            \caption*{(b) Real Dataset}
            \label{fig:real}
        \end{minipage} \\
    \end{tabular}
    \caption{\textbf{Results of Flare-occupancy Mask Generation network (FMG):} on (a) synthetic datasets and (b) real datasets. Corresponding manually annotated masks for the real dataset in (b) match with the learned masks generated by our FMG.}
    \label{fig:flaremask}
\end{figure}

We use PSPNet \cite{zhao2017pyramid}, a semantic segmentation model, to generate the flare occupancy mask. 
We captured 80 different real flare patterns in a dark setting by showing torch light in multiple directions. We manually annotated these flare patterns to get binary flare-occupancy masks to highlight flare and flare-free regions. These flare patterns are imposed on the 24K Flickr image dataset\cite{zhang2018single} with a random affine transform on each flare image following the Flare7K methodology and the same affine transform is applied to annotated flare-occupancy mask to generated corresponding ground-truth flare-occupancy mask. This random affine transform simulates diverse flare patterns in multiple directions at various spatial locations. We used this dataset : flare images with their corresponding flare-occupancy masks to train PSPNet. We also observed that the inherent class (two classes : 1) regions w/ flare, 2) regions w/o flare) imbalance problem which further hinders the model's performance: regions affected by flare in an image is far less compared to flare-free regions, which creates class imbalance issues during the training process. So, we empirically assigned weights in the ratio 5:1 to the respective classes in calculating the binary cross-entropy loss during training with other settings unaltered. With these modifications, we are able to generate flare-occupancy mask to detect the flare regions considerably well on both synthetic images with quantitative metrics of 0.81/0.94 (mIoU/mAcc) and on real images, which is evident from the results of our proposed FMG module on both the synthetic and real datasets in Fig.~\ref{fig:flaremask}.

\vspace{-1em}

\vspace{-0.75em}
\subsection{View and Point Sampling Modules}
 \label{sec:viewandpoint}
 We proposed two sampling modules to effectively sample source, target views and points on each ray while rendering each target view to accurately sample the information from only flare-free regions to enhance scene representation. 
 \vspace{-0.75em}
\subsubsection{View Sampler}
 \label{sec:viewsampler}
As seen from our captured real sample dataset, Fig.~\ref{fig:flaremask}, some images are highly corrupted with flare. Utilizing any of these source images for rendering a target view introduces flare artifacts, hence compromising the scene representation. Therefore, we adopt a strategy of selecting nearby source views that are less impacted by flare using flare occupancy mask as a guide. Our sampling strategy  1) We exclude the image as a target view from training, if its flare occupancy(\% of flare effected region) is beyond $10\%$, so this view can be rendered as a novel view during inference. 2) For rendering each target view, we only sample the nearby source views which are less effected by flare using flare occupancy information from flare occupancy mask. These two view sampling strategies ensure that network weights are not influenced by the largely effected flare views during optimization.
\vspace{-0.75em}
\subsubsection{Point Sampler} 
\label{sec:pointsampler}
While rendering ray, we sample points from corresponding epipolar lines of sampled source views. These sample points which fall on flare effected region will sample flare, hence it corrupts the target view while rendering. To nullify the effect of these flare sample points, we make the attention values due to these sample points zero in view transformer. This technique ensures that we sample almost no information from flare affected regions. Hence, scene representation by GN-FR is mostly unaffected by the flare. Our updated attention in view transformer is a simple multiplication with mask. 
\begin{align}
    \begin{split}
        A^{\prime} & = A * (1-M) \\
    \end{split}
\end{align}


\vspace{-0.75em}
\subsection{Flare Occupancy Mask Guided Loss}
 \label{sec:loss}
Capturing a real dataset with and without flare is highly impractical to train our GN-FR module, we leverage an unsupervised training strategy (we don't have ground-truth image w/o flare) by constructing flare occupancy mask guided loss, where we multiply the flare-occupancy mask element wise to both the target image and its corresponding rendered image and then apply MSE loss; this ensures that the regions w/ flare in target image are rendered from corresponding regions w/o flare from the source views as like a novel region of the target view. Our idea follows as NeRF renders a novel views, it can also render novel regions in the views. This presupposes that at least one of the source views remains unaffected by flare in the target region while acknowledging flare presence in the specified area. As flare is view-dependent, this condition is mostly valid; a portion of the scene is unaffected by flare, at least in one of the images.

\begin{equation}
    \boldsymbol{L_{unsup}} = \|Pred \odot (1-M) - Target \odot (1-M)\|
\end{equation}

\vspace{-0.75em}
\section{Experiments}
\vspace{-0.75em}
\subsection{Settings}
Following GNT, we construct a dictionary of each target view with its sampled source views for training by first selecting a target view, and then identifying a pool of $k \times N$ nearby views, from which N views are sampled as source views following the View Sampler strategy (Sec.~\ref{sec:viewsampler}). This sampling strategy ensures very minimal information from flare regions is passed through the network while rendering a target view. During training, the values for k and N are uniformly sampled at random from (1, 3) and (8, 12) respectively. From these views, N views are sampled at the end following the View Sampler strategy.
\vspace{-0.75em}
\subsubsection{Our Proposed Flare Dataset}

\begin{figure}[h]
    \centering
    \begin{tabular}{cc}
        \begin{minipage}{0.49\textwidth}
            \centering
            \includegraphics[width=\textwidth]{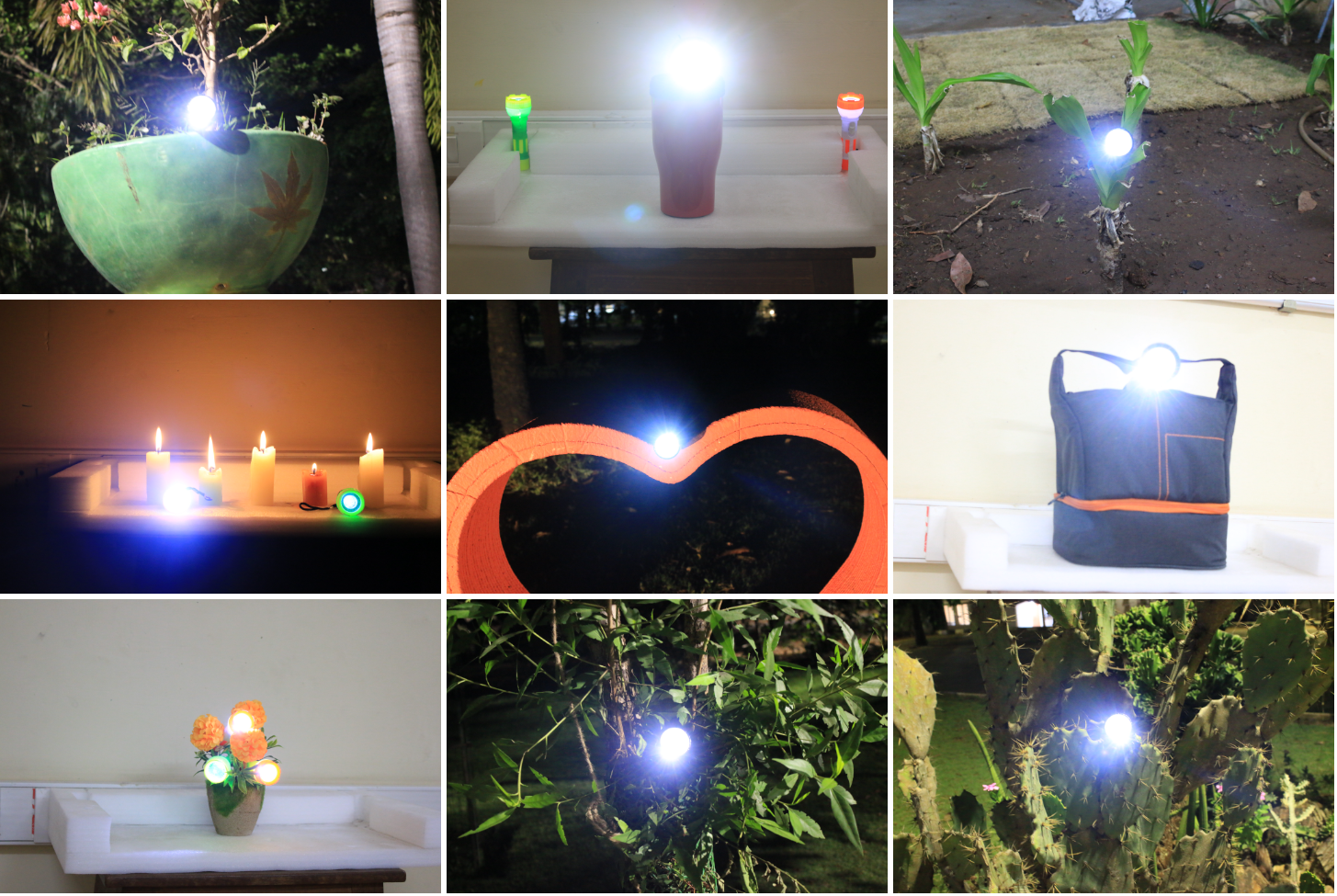}
            \caption*{(a) Our Captured Flare Dataset}
            \label{fig:captured_dataset}
        \end{minipage} &
        \begin{minipage}{0.49\textwidth}
            \centering
            \includegraphics[width=\textwidth]{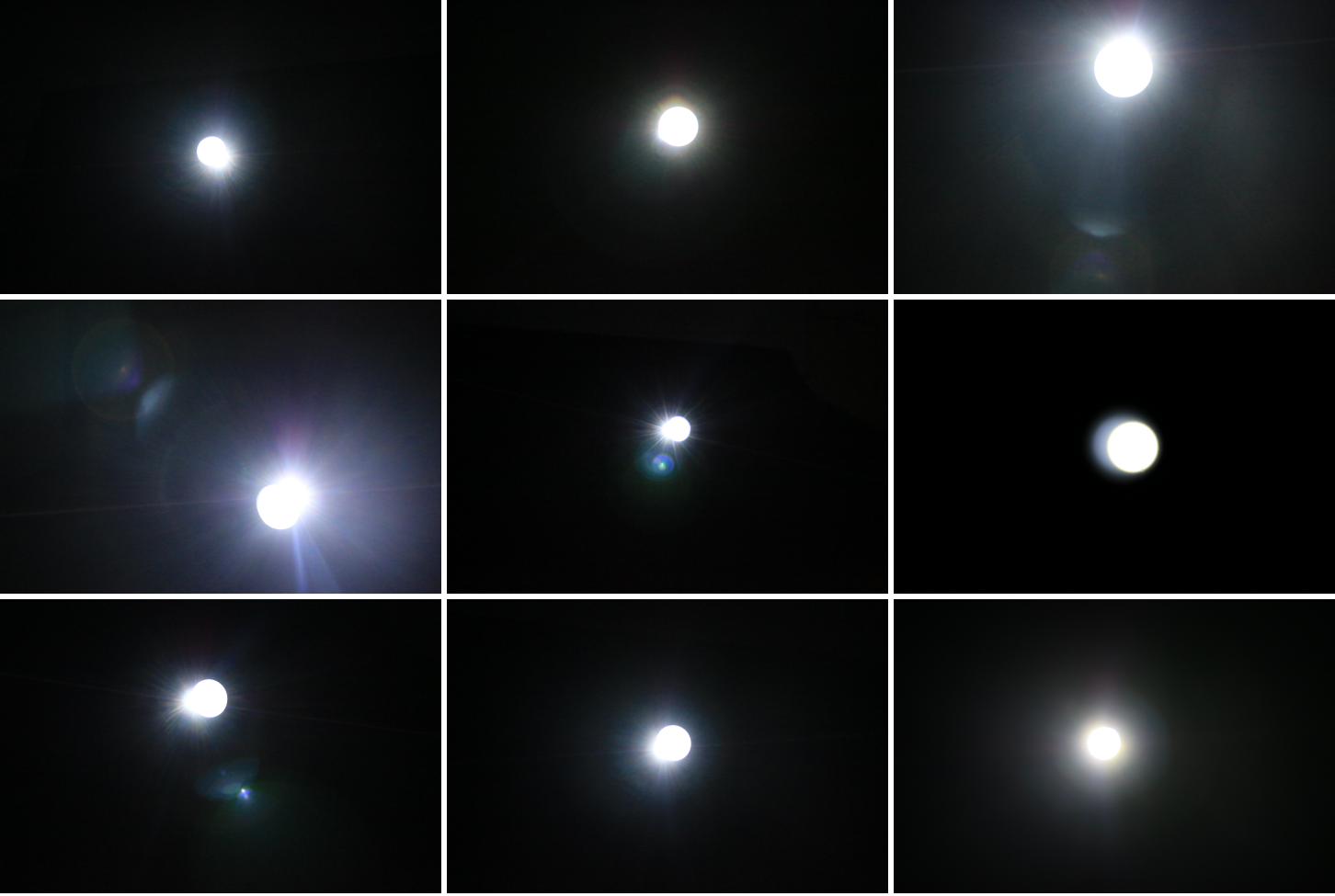}
            \caption*{(b) Our Captured Flare Patterns}
            \label{fig:flare_pattern_dataset}
        \end{minipage} \\
    \end{tabular}
    \caption{(a) Samples from our captured real flare-3D dataset of 17 scenes and (b) Samples from our captured real flare pattern dataset of 80 images, consisting of diverse flare patterns (both reflective and scattering).}
    \label{fig:flaredataset}
\end{figure}

We captured 80 different flare patterns with a light source in a dark setting as shown in Fig.~\ref{fig:flaredataset} and then imposed these real flares on the 24K Flickr dataset with a random affine transform on each flare pattern image, this account for diverse flare patterns in multiple directions and orientations. Also, we collected a real flare 3D dataset of 17 scenes, comprising of 782 images to validate our method on real dataset. These are manually annotated with MATLAB using image segmentation application to generate flare occupancy mask. Please refer to the Fig.~\ref{fig:flaremask} to the samples from our dataset.
\vspace{-0.75em}
\subsubsection{Training/Inference Details}
We train the FMG following the default settings of PSPNet with modified weights of [5,1] (Sec.~\ref{sec:fmg}) to binary cross-entropy loss function. GN-FR is trained an end-to-end fashion on generated flare dataset with corresponding masks from 44 scenes of IBRNet dataset using the Adam optimizer to minimize the mean-squared error between predicted and target RGB pixel values using masking loss. For all our experiments, we train for 500,000 iterations with 256 rays sampled in each iteration with initialization from GNT. To test on real scenes, for better results our model is finetuned for 50k iterations. We run all our experiments on NVIDIA RTX 3090, 24GB. Unless otherwise stated, other settings of our model remains the same as that of GNT.
\vspace{-1em}
\subsection{Quantitative Evaluations}
\label{quantitative}
For quantitative evaluation, as real dataset w/ flares and w/o flares is not available, we took IBRNet dataset, synthetically imposed the flares in random fashion and then send those to our network. As ours is the first work which does flare removal using NeRF, we compare our work with two methods, one is scene specific which is vanilla NeRF\cite{mildenhall2020nerf} and the other is generalizable one, which is GNT. We first de-flare the images using Flare7K++ (existing SOTA flare removal technique) and then sent them to both NeRF and GNT for proper comparison. As we see from the Table~\ref{tab:realscenes}, our method is producing superior results in all the metrics, surprisingly having 4dB PSNR improvement over the second best method.
\begin{table}[th]
  \centering
  \caption{
  Quantitative results on  real scenes with synthetically imposed flare. The \sethlcolor{black!30}\hl{best} scores and \sethlcolor{black!15}\hl{second best} scores are highlighted with their respective colors. 
  }
  \vspace{1em}
  \resizebox{0.75\columnwidth}{!}{
  \begin{tabular}{lcccc}
  
    \toprule
    Models & \makecell{Cross-scene \\ Generalization} & \hspace{0.1em}PSNR$\uparrow$\hspace{0.1em} & SSIM$\uparrow$\hspace{0.1em} & LPIPS$\downarrow$\hspace{0.1em}\\
    \midrule
    NeRF\cite{mildenhall2020nerf}& \xmark & 21.92 & 0.827 & 0.045\\
    \midrule
    GNT\cite{gnt} & \cmark & \cellcolor{black!15}22.44 & 0.867 & \cellcolor{black!15}0.0386 \\ 
    \midrule
    (Flare7K++)\cite{dai2023flare7k++}+(NeRF)\cite{mildenhall2020nerf}& \xmark & 18.63 & 0.769 & 0.052\\
    \midrule
    (Flare7K++)\cite{dai2023flare7k++}+(GNT)\cite{gnt} & \cmark & 18.85 &  0.812 & 0.056 \\ 
    \midrule
    Ours & \cmark & \cellcolor{black!30}26.18 & \cellcolor{black!30}0.8815 & \cellcolor{black!30}0.034 \\ 
    \bottomrule
  \end{tabular}
  }
  
  \label{tab:realscenes}
  \vspace{-1em}
\end{table}

\vspace{-0.75em}
\subsection{Qualitative Results}
\begin{figure*}[!ht]
  \centering
  \includegraphics[width=0.7\textwidth]{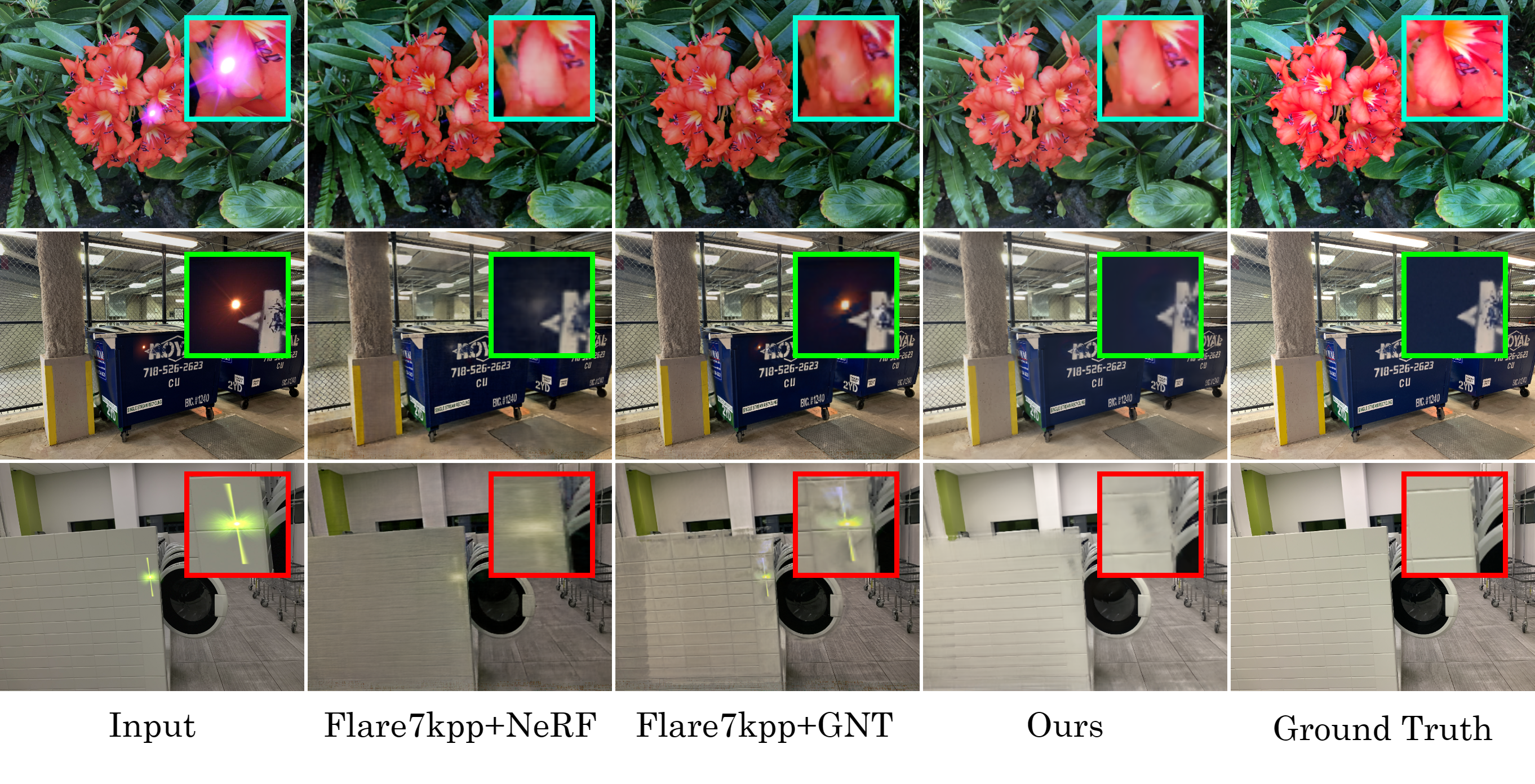}
   \hspace{0.75em}
  \caption{\textbf{Qualitative results on Synthetic dataset:} GN-FR framework is able to consistently remove the diverse flares present in the synthetically imposed flare images. As the insets highlight, GN-FR surpasses other methods in removing flare artifacts.}
  \label{fig:synthetic_qual}
\end{figure*}

\begin{figure*}[!ht]
    \centering
  \includegraphics[width=0.7\textwidth]{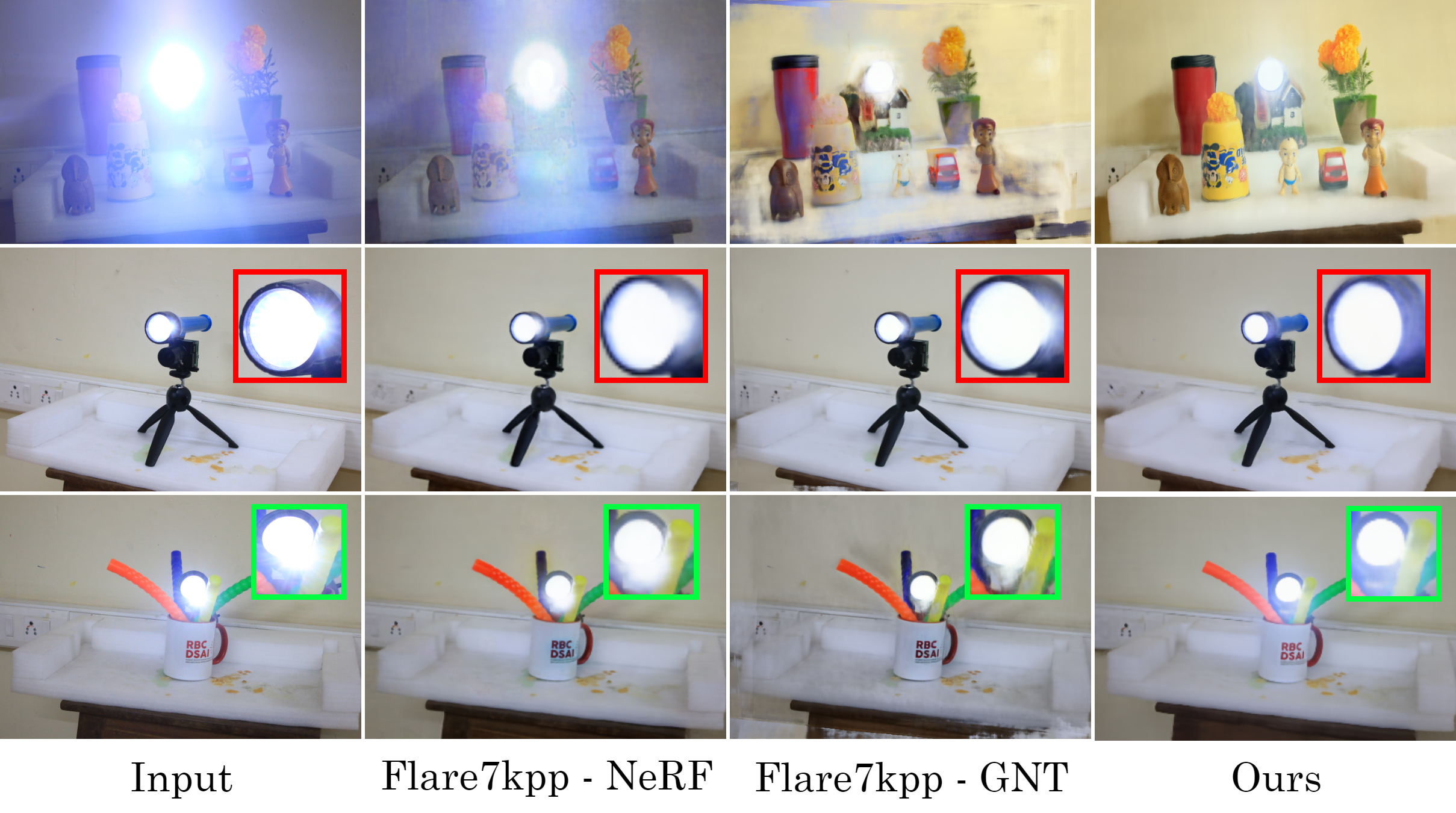}
  \caption{\textbf{Qualitative results on Real dataset:} Our GN-FR framework is able to remove the flare artifacts from flare-corrupted images effectively. It is able to render more visually appealing results consistently compared to other methods.}
  \label{fig:real_qualitative}

\end{figure*}

\begin{figure*}[!ht]
    \centering
  \includegraphics[width=0.8\textwidth]{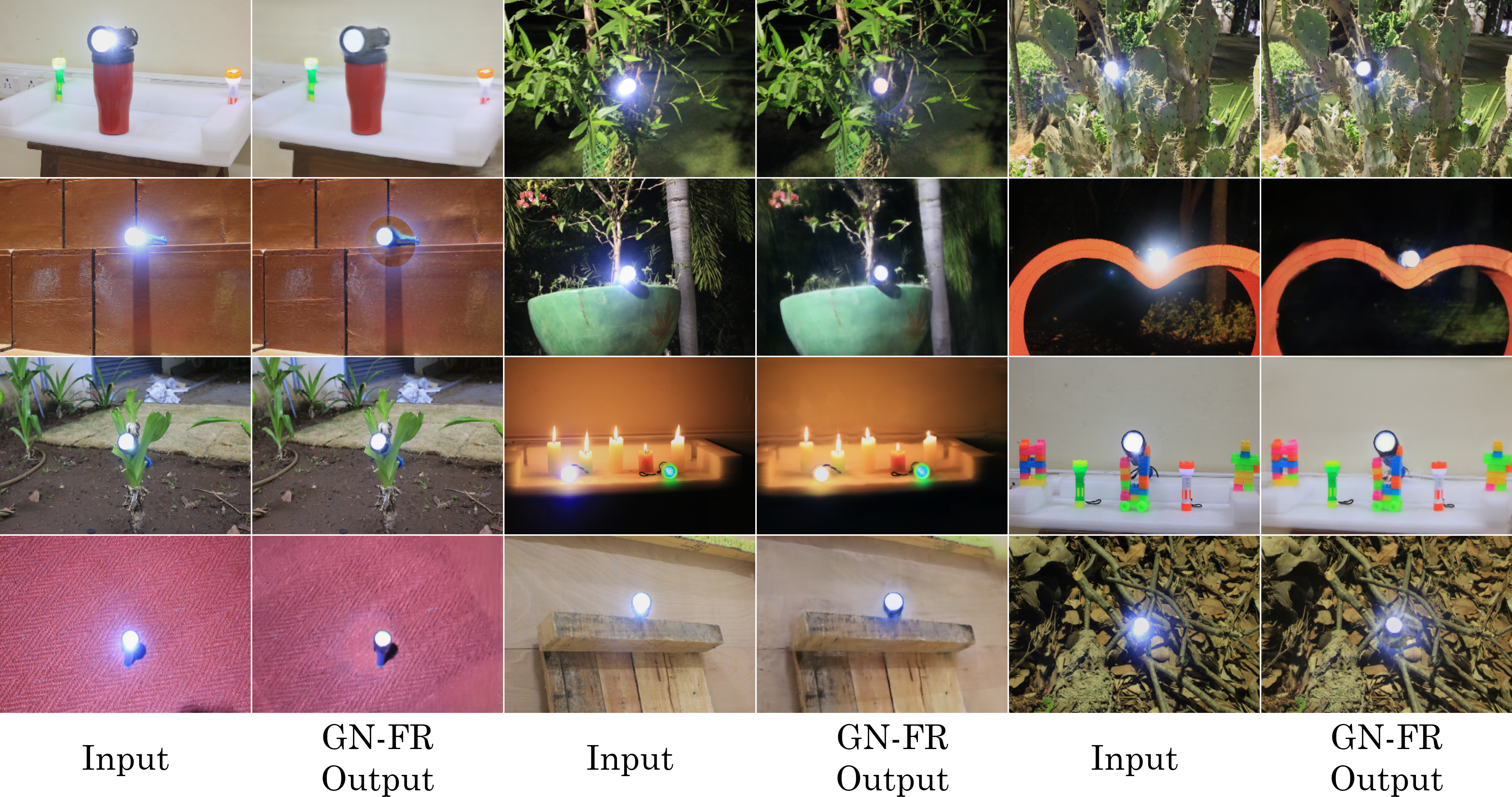}
  \caption{\textbf{Flare removal results on captured Real dataset:} GN-FR is able to consistently remove the flare-artifacts on different scenes of our real flare-3D dataset and also reproduce the lost scene information covered up due to flare effects(Zoom-in for better visualization). }
  \label{fig:supply_results}
\end{figure*}

For qualitative evaluation, we tested our method on both synthetic and captured real datasets with methods (Flare7K++)+(GNT) and (Flare7K++)+(NeRF). As we see from the Figs.~\ref{fig:synthetic_qual} and ~\ref{fig:real_qualitative}, our method is working well both on synthetically imposed flare scenes as well on real scenes. Whereas other methods are having difficulty in removing flare consistently, either they leave artifacts or introduces artifacts in flare-free regions. As reported in Fig.~\ref{fig:supply_results} our method is able to remove the flare artifacts and restore the lost scene content effected due to flare very well on the captured real dataset. For more results on synthetic as well real datasets, please refer supplementary material.
\vspace{-1em}

\vspace{-1em}

\vspace{3em}

\subsection{Ablation Study}
\label{sec:ablation}
\begin{figure*}[!ht]
    \centering
  \includegraphics[width=0.7\textwidth]{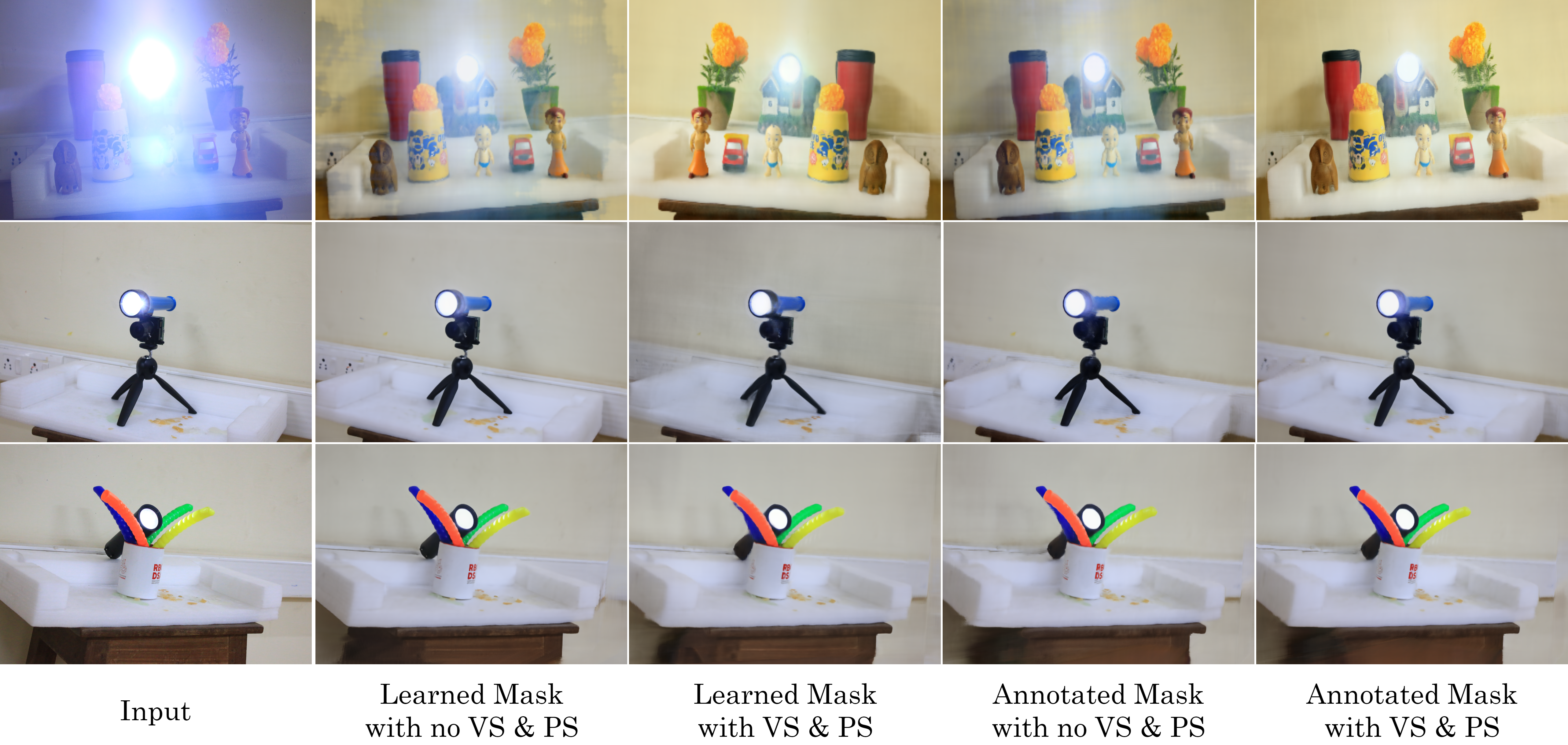}
  \caption{\textbf{Effectiveness of Learned mask together with PS and VS modules:} Proposed modules(PS \& VS) contribute positively to render the target views with very minimal flare artifacts. Rendered target views with learned masks are almost on par with annotated masks.}
  \label{fig:real_qual}
\end{figure*}

\textbf{Without View Sampler (VS), Point Sampler (PS) modules and generated masks:} 
Our contributions include masking loss, VS and PS modules and flare occupancy masking network. So we did the ablations without VS and PS modules, as we can see from Fig.~\ref{fig:real_qual}, inclusion of PS and VS modules contribute to effective removal of flare artifacts, where flare is significant. As we see from the results, our method including all gives the SOTA results. Also, render results with generated flare occupancy masks are on on par with annotated masks, Fig.~\ref{fig:real_qual}.

\vspace{-0.75em}
\section{Conclusion}
We present GN-FR, an innovative unsupervised neural rendering framework designed to effectively remove flares by leveraging multi-view flare-free information from neighboring source views. Our approach minimizes flare artifacts in the target view, achieving superior performance. We collected real flare pattern dataset to generate flare occupancy mask which will attend effectively to the flare-free regions while rendering. Also, we contribute a 3D flare dataset with ground-truth flare occupancy mask annotations. Our model demonstrates significant effectiveness in flare removal across both synthetic and real-world scenarios, setting a new state-of-the-art in both quantitative and qualitative assessments. To our knowledge, GN-FR is the first framework to utilize multi-view information for enhanced flare removal, opening new avenues for research in this field. Future work will aim to extend our framework to handle diverse flare types and explore zero-shot generalization on real-world scenes.
\\
\\
\textbf{Acknowledgement:} 
 This work was supported in part by IITM Pravartak Technologies Foundation.

\begin{thebibliography}{23}
\providecommand{\natexlab}[1]{#1}
\providecommand{\url}[1]{\texttt{#1}}
\expandafter\ifx\csname urlstyle\endcsname\relax
  \providecommand{\doi}[1]{doi: #1}\else
  \providecommand{\doi}{doi: \begingroup \urlstyle{rm}\Url}\fi

\bibitem[Barron et~al.(2021)Barron, Mildenhall, Tancik, Hedman, Martin-Brualla, and Srinivasan]{Barron_2021_ICCV}
Jonathan~T. Barron, Ben Mildenhall, Matthew Tancik, Peter Hedman, Ricardo Martin-Brualla, and Pratul~P. Srinivasan.
\newblock Mip-nerf: A multiscale representation for anti-aliasing neural radiance fields.
\newblock In \emph{Proceedings of the IEEE/CVF International Conference on Computer Vision (ICCV)}, pages 5855--5864, October 2021.

\bibitem[Chen et~al.(2021{\natexlab{a}})Chen, Xu, Zhao, Zhang, Xiang, Yu, and Su]{chen2021mvsnerf}
Anpei Chen, Zexiang Xu, Fuqiang Zhao, Xiaoshuai Zhang, Fanbo Xiang, Jingyi Yu, and Hao Su.
\newblock Mvsnerf: Fast generalizable radiance field reconstruction from multi-view stereo.
\newblock In \emph{Proceedings of the IEEE/CVF international conference on computer vision}, pages 14124--14133, 2021{\natexlab{a}}.

\bibitem[Chen et~al.(2021{\natexlab{b}})Chen, He, Wang, Ren, Lim, and Shrivastava]{chen2021nerv}
Hao Chen, Bo~He, Hanyu Wang, Yixuan Ren, Ser-Nam Lim, and Abhinav Shrivastava.
\newblock Ne{RV}: Neural representations for videos.
\newblock In \emph{NeurIPS}, 2021{\natexlab{b}}.

\bibitem[Dai et~al.(2023{\natexlab{a}})Dai, Li, Zhou, Feng, Luo, and Loy]{dai2023flare7k++}
Yuekun Dai, Chongyi Li, Shangchen Zhou, Ruicheng Feng, Yihang Luo, and Chen~Change Loy.
\newblock Flare7k++: Mixing synthetic and real datasets for nighttime flare removal and beyond.
\newblock \emph{arXiv preprint arXiv:2306.04236}, 2023{\natexlab{a}}.

\bibitem[Dai et~al.(2023{\natexlab{b}})Dai, Luo, Zhou, Li, and Loy]{dai2023nighttime}
Yuekun Dai, Yihang Luo, Shangchen Zhou, Chongyi Li, and Chen~Change Loy.
\newblock Nighttime smartphone reflective flare removal using optical center symmetry prior.
\newblock In \emph{Proceedings of the IEEE/CVF Conference on Computer Vision and Pattern Recognition}, pages 20783--20791, 2023{\natexlab{b}}.

\bibitem[Dosovitskiy et~al.(2020)Dosovitskiy, Beyer, Kolesnikov, Weissenborn, Zhai, Unterthiner, Dehghani, Minderer, Heigold, Gelly, et~al.]{dosovitskiy2020image}
Alexey Dosovitskiy, Lucas Beyer, Alexander Kolesnikov, Dirk Weissenborn, Xiaohua Zhai, Thomas Unterthiner, Mostafa Dehghani, Matthias Minderer, Georg Heigold, Sylvain Gelly, et~al.
\newblock An image is worth 16x16 words: Transformers for image recognition at scale.
\newblock \emph{arXiv preprint arXiv:2010.11929}, 2020.

\bibitem[Hu et~al.(2022)Hu, Liu, Chen, Shen, and Jia]{hu2022efficientnerf}
Tao Hu, Shu Liu, Yilun Chen, Tiancheng Shen, and Jiaya Jia.
\newblock Efficientnerf efficient neural radiance fields.
\newblock In \emph{Proceedings of the IEEE/CVF Conference on Computer Vision and Pattern Recognition}, pages 12902--12911, 2022.

\bibitem[Kotp and Torki(2024)]{kotp2024flare}
Yousef Kotp and Marwan Torki.
\newblock Flare-free vision: Empowering uformer with depth insights.
\newblock In \emph{ICASSP 2024-2024 IEEE International Conference on Acoustics, Speech and Signal Processing (ICASSP)}, pages 2565--2569. IEEE, 2024.

\bibitem[Liu et~al.(2022)Liu, Peng, Liu, Wang, Wang, Theobalt, Zhou, and Wang]{liu2022neural}
Yuan Liu, Sida Peng, Lingjie Liu, Qianqian Wang, Peng Wang, Christian Theobalt, Xiaowei Zhou, and Wenping Wang.
\newblock Neural rays for occlusion-aware image-based rendering.
\newblock In \emph{Proceedings of the IEEE/CVF Conference on Computer Vision and Pattern Recognition}, pages 7824--7833, 2022.

\bibitem[Mildenhall et~al.(2020)Mildenhall, Srinivasan, Tancik, Barron, Ramamoorthi, and Ng]{mildenhall2020nerf}
Ben Mildenhall, Pratul~P. Srinivasan, Matthew Tancik, Jonathan~T. Barron, Ravi Ramamoorthi, and Ren Ng.
\newblock Nerf: Representing scenes as neural radiance fields for view synthesis.
\newblock In \emph{ECCV}, 2020.

\bibitem[Mildenhall et~al.(2022)Mildenhall, Hedman, Martin-Brualla, Srinivasan, and Barron]{mildenhall2022nerf}
Ben Mildenhall, Peter Hedman, Ricardo Martin-Brualla, Pratul~P Srinivasan, and Jonathan~T Barron.
\newblock Nerf in the dark: High dynamic range view synthesis from noisy raw images.
\newblock In \emph{Proceedings of the IEEE/CVF Conference on Computer Vision and Pattern Recognition}, pages 16190--16199, 2022.

\bibitem[Qiao et~al.(2021)Qiao, Hancke, and Lau]{qiao2021light}
Xiaotian Qiao, Gerhard~P Hancke, and Rynson~WH Lau.
\newblock Light source guided single-image flare removal from unpaired data.
\newblock In \emph{Proceedings of the IEEE/CVF International Conference on Computer Vision}, pages 4177--4185, 2021.

\bibitem[T et~al.(2023)T, Wang, Chen, Chen, Venugopalan, and Wang]{gnt}
Mukund~Varma T, Peihao Wang, Xuxi Chen, Tianlong Chen, Subhashini Venugopalan, and Zhangyang Wang.
\newblock Is attention all that ne{RF} needs?
\newblock In \emph{The Eleventh International Conference on Learning Representations}, 2023.
\newblock URL \url{https://openreview.net/forum?id=xE-LtsE-xx}.

\bibitem[Verbin et~al.(2022)Verbin, Hedman, Mildenhall, Zickler, Barron, and Srinivasan]{verbin2022ref}
Dor Verbin, Peter Hedman, Ben Mildenhall, Todd Zickler, Jonathan~T Barron, and Pratul~P Srinivasan.
\newblock Ref-nerf: Structured view-dependent appearance for neural radiance fields.
\newblock In \emph{2022 IEEE/CVF Conference on Computer Vision and Pattern Recognition (CVPR)}, pages 5481--5490. IEEE, 2022.

\bibitem[Wang et~al.(2022{\natexlab{a}})Wang, Cui, Salcudean, and Wang]{wang2022generalizable}
Dan Wang, Xinrui Cui, Septimiu Salcudean, and Z~Jane Wang.
\newblock Generalizable neural radiance fields for novel view synthesis with transformer.
\newblock \emph{arXiv preprint arXiv:2206.05375}, 2022{\natexlab{a}}.

\bibitem[Wang et~al.(2021)Wang, Wang, Genova, Srinivasan, Zhou, Barron, Martin-Brualla, Snavely, and Funkhouser]{wang2021ibrnet}
Qianqian Wang, Zhicheng Wang, Kyle Genova, Pratul~P Srinivasan, Howard Zhou, Jonathan~T Barron, Ricardo Martin-Brualla, Noah Snavely, and Thomas Funkhouser.
\newblock Ibrnet: Learning multi-view image-based rendering.
\newblock In \emph{Proceedings of the IEEE/CVF Conference on Computer Vision and Pattern Recognition}, pages 4690--4699, 2021.

\bibitem[Wang et~al.(2022{\natexlab{b}})Wang, Cun, Bao, Zhou, Liu, and Li]{wang2022uformer}
Zhendong Wang, Xiaodong Cun, Jianmin Bao, Wengang Zhou, Jianzhuang Liu, and Houqiang Li.
\newblock Uformer: A general u-shaped transformer for image restoration.
\newblock In \emph{Proceedings of the IEEE/CVF conference on computer vision and pattern recognition}, pages 17683--17693, 2022{\natexlab{b}}.

\bibitem[Wu et~al.(2021)Wu, He, Xue, Garg, Chen, Veeraraghavan, and Barron]{wu2021train}
Yicheng Wu, Qiurui He, Tianfan Xue, Rahul Garg, Jiawen Chen, Ashok Veeraraghavan, and Jonathan~T Barron.
\newblock How to train neural networks for flare removal.
\newblock In \emph{Proceedings of the IEEE/CVF International Conference on Computer Vision}, pages 2239--2247, 2021.

\bibitem[Yu et~al.(2021)Yu, Ye, Tancik, and Kanazawa]{yu2021pixelnerf}
Alex Yu, Vickie Ye, Matthew Tancik, and Angjoo Kanazawa.
\newblock pixelnerf: Neural radiance fields from one or few images.
\newblock In \emph{Proceedings of the IEEE/CVF Conference on Computer Vision and Pattern Recognition}, pages 4578--4587, 2021.

\bibitem[Zhang et~al.(2023)Zhang, Ouyang, Liu, Wang, Kong, and Jin]{zhang2023ff}
Dafeng Zhang, Jia Ouyang, Guanqun Liu, Xiaobing Wang, Xiangyu Kong, and Zhezhu Jin.
\newblock Ff-former: Swin fourier transformer for nighttime flare removal.
\newblock In \emph{Proceedings of the IEEE/CVF Conference on Computer Vision and Pattern Recognition}, pages 2824--2832, 2023.

\bibitem[Zhang et~al.(2018)Zhang, Ng, and Chen]{zhang2018single}
Xuaner Zhang, Ren Ng, and Qifeng Chen.
\newblock Single image reflection separation with perceptual losses.
\newblock In \emph{Proceedings of the IEEE conference on computer vision and pattern recognition}, pages 4786--4794, 2018.

\bibitem[Zhao et~al.(2017)Zhao, Shi, Qi, Wang, and Jia]{zhao2017pyramid}
Hengshuang Zhao, Jianping Shi, Xiaojuan Qi, Xiaogang Wang, and Jiaya Jia.
\newblock Pyramid scene parsing network.
\newblock In \emph{Proceedings of the IEEE conference on computer vision and pattern recognition}, pages 2881--2890, 2017.

\bibitem[Zhou et~al.(2023)Zhou, Liang, Chen, Huang, Yang, and Li]{zhou2023improving}
Yuyan Zhou, Dong Liang, Songcan Chen, Sheng-Jun Huang, Shuo Yang, and Chongyi Li.
\newblock Improving lens flare removal with general-purpose pipeline and multiple light sources recovery.
\newblock In \emph{Proceedings of the IEEE/CVF International Conference on Computer Vision}, pages 12969--12979, 2023.

\end{thebibliography}

\end{document}